\documentclass[authoryear,preprint,12pt]{elsarticle}

\usepackage{amssymb}

\journal{arXiv}

\begin{document}

\begin{frontmatter}


 \author{Georgios Mastorakis}
\ead{g.mastorakis@ieee.org}
 \address{CogniLabs Ltd\fnref{label3}}

\title{Human-like machine learning: limitations and suggestions}




\begin{abstract}
This paper attempts to address the issues of machine learning in its current implementation. It is known that machine learning algorithms require a significant amount of data for training purposes, whereas recent developments in deep learning have increased this requirement dramatically. The performance of an algorithm depends on the quality of data and hence, algorithms are as good as the data they are trained on. Supervised learning is developed based on human learning processes by analysing named (i.e. annotated) objects, scenes and actions. Whether training on large quantities of data (i.e. big data) is the right or the wrong approach, is debatable. The fact is, that training algorithms the same way we learn ourselves, comes with limitations. This paper discusses the issues around applying a human-like approach to train algorithms and the implications of this approach when using limited data. Several current studies involving non-data-driven algorithms and natural examples are also discussed and certain alternative approaches are suggested.

\end{abstract}


\end{frontmatter}


\section{Introduction}
\label{intro}

The human learning process involves the gradual brain development as we grow from infancy towards adulthood. A machine however, does not share this evolutionary process and complies with its initial extended capacity which may outperform our capabilities on particular learning tasks \cite{mldif}. This capability though, may not always provide accurate results and can often lead to classification errors e.g  a clementine may be confused with an orange. 
The primitive humans' ability to process learning involved  a type of a rather incomplete ``unsupervised learning", as they observed objects but had not developed a language system to annotate them. The cognitive process where the next generations classified or named objects, is effectively what unsupervised machine learning does by clustering. The unsupervised approach can separate classes, nevertheless, an algorithm has no actual understanding of the action or object that is clustered. On the other hand, in supervised approach, every object or action has an annotation, hence the training algorithm uses an identifier of this particular action or object. In both learning methods the algorithms are fed with - what we think is - significant, as we select the training images or videos. Sometimes we may also choose to omit data of poor visual quality (i.e. blurred or saturated images, etc.). Effectively, the learning approach is controlled by human decision.

Currently, successful data-driven approaches assume that algorithms would work optimally if they are trained on data which covers all possible types of representative examples of actions or objects. However, such complete data is difficult to capture due to a number of limiting factors i.e. risks, privacy etc. On the other hand, the human brain functions in a control-free manner where visual, acoustic, thermal and other senses are processed and stored for future recognition tasks and therefore, the information is unconstrained. Furthermore, the human brain develops and performs self-learning behaviours/tasks by using physical functions (i.e.using muscles) for activities such as standing or walking. Learning how to walk, throw a ball or lying down, are self-taught actions and parametrised by physics' dynamics. However, these actions require further self-correction and self-learning in order to walk stably or throw the ball accurately, a task that may require long term practising. 

Another human ability is to recognise objects or actions which are partially occluded. For example, a human would easily recognise an occluded person walking behind a low fence covering 70\% of their height, or identify a banana only by its stem. However in computer vision, the issue around recognising occluded objects or actions has not been adequately addressed or studied, with the majority of studies focusing primarily on identifying fully visible events or objects.




The two main issues discussed in this paper are related to the machine learning algorithms' ability to recognise objects or events based on limited training data and whether this training data is indeed representative. One approach to overcome the issue around data availability is the use of synthetic data \cite{gaidon2018reasonable} created by researchers using gaming and simulation engines. Such data is mainly introduced to increase the number of training examples. However, these datasets do not necessarily capture difficult to replicate in real life incidents nor do they include missing event data (i.e. car crashes, human falls). One of the questions raised is whether data-driven approaches are required to model an action, event or behaviour of humans or objects. More specifically, do we require the algorithm to have pre-knowledge of the event's behaviour in order to perform robust action recognition? There are only a few examples available in the literature that use simulation instead of recorded data  to model an activity such as: walking \cite{brubaker2010physics} by developing a bipedal physics simulation, object falling \cite{li2016fall} by using synthetic blocks, and human falling \cite{mastorakis2018fall} by using a myoskleletal simulation. However, these approaches are sparse and limited to a particular recognition task, partially due to the fact that utilising simulation approaches require mathematical complexity for their development. An alternative use of a simulation model is seen in \cite{heess2017emergence} where authors utilise a stick model to perform tasks such as running (i.e. reinforcement learning). These approaches will be further discussed later in this paper.

The above issues around human-like/human-guided machine learning and lack of data availability highlight the difficulties within data-driven algorithmic performance and hence, alternative methods need to be identified. Artificial Intelligence algorithms could enhance their capabilities if the effect of physics or other parameters on a particular event were comprehended and a more holistic level of awareness was achieved. This integrated approach would not require the use of data covering all possible examples of an action -which is a complication in most learning algorithms- and as a result it would enhance the performance of a learning system, utilising additional non-data driven parameters. This paper focuses on data availability issues around human action recognition that are linked to poor performance of machine learning algorithms. It also refers to approaches using simulation and self-learning capabilities. It is organised as follows: i) a detailed discussion of the issues of current available data, ii) exploration of the issues of machine learning, (iii) review and discussion of the simulation and non-data-driven approaches, (iv) identifying natural examples, followed by (v) a discussion and (vi) conclusion.


\section{Dataset issues}
\label{bigdataissues}

The requirement of using more data has increased during the past few years as it is also highlighted by the evidence described in \cite{zhu2016we} where it shows that using more data leads to better algorithmic performance. Other studies however, suggest the opposite (i.e. less is more) for object detection \cite{zhu2012we} or are inconclusive as to whether more data is beneficial e.g for facial action recognition \cite{girard2015much}. Others also suggest the merging of existing datasets in order to tackle the data availability issue \cite{schuldhaus2014towards}.

Several datasets for action recognition were developed as seen in several review studies \cite{zhang2016rgb, firman2016rgbd, singh2018video}. These datasets were created by researchers to help them develop and evaluate algorithms. Generally, these datasets represent the actions in such a way that the analysis is performed using\textit{ cleaner} data than when the similar action is observed in the wild. Such data is recorded and annotated prior to being processed by algorithms.

As previously discussed, machine learning algorithms inevitably require a significant amount of data for training. This is however a limitation in data recording of actions that involve a level of risk. Therefore, these datasets are sparse and of questionable quality in terms of how realistically these events are performed. An example of these events which involve risk, is datasets for fall or aggression detection - events that do not occur often but rather accidentally- and as a result we are not prepared or willing to perform such an action for the sake of data recording alone. Even when we decide to perform such staged actions, we may be reluctant to act realistically, due to the risk of injury. In \cite{mabrouk2018abnormal}, authors review the current datasets for violence detection which are derived from sport events (i.e. ice-hockey), YouTube videos and CCTV footage. Nevertheless, the size of data is limited with less than 800 video samples across 6 datasets. Additionally, the fall event datasets as seen in Table \ref{table:publicDatasets} specify the number of subjects and samples of each dataset. It is observed that data availability is also an issue for this type of action, given the limited number of fall samples. In the above-discussed datasets for action recognition, falls, violent or risky actions, constitute a very small class of data samples in comparison to less risky actions, such as walking, sitting, greeting etc.

\begin{table}[]
\centering
\scriptsize
\caption{Datasets of visual fall data. R: RGB data, IR: infrared data, D: depth data, A: accelerometer data, S: Kinect skeleton data. The table shows the different fall event datasets of several sensor technologies. Noticeable is the number of fall events if compared with the ADLs as well as how small the fall number is in general}
\label{table:publicDatasets}
\begin{tabular}{c|c|c|c|c|c}
Dataset               & Subjects & Actions & \begin{tabular}[c]{@{}c@{}}Fall\\ Samples\end{tabular} & \begin{tabular}[c]{@{}c@{}}ADL\\ Samples\end{tabular} & \begin{tabular}[c]{@{}c@{}}Data\\ Type\end{tabular} \\
\hline
Multiple cameras~\cite{auvinet2011fall} & 1        & 9       & 24                                                     & 99                                                    &  R                                                 \\
LE2i~\cite{charfi2013optimized}              & 9        & 7       & 143                                                    & 48                                                    &  R                                                 \\
TST v2~\cite{gasparrini2016proposal}                & 11       & 5       &                                                        &                                                       &  D, S, A                                             \\
UR~\cite{kwolek2014human}               & 5        & 6       & 30                                                     & 40                                                    &  R, D, A                                           \\
SDUFall~\cite{ma2014depth}             & 20       & 6       & 200                                                    & 1000                                                  & R, D, S                                           \\
Fall Detection~\cite{zhang2012viewpoint}      & 6        & 8       & 26                                                     & 61                                                    & D                                                   \\
EDF~\cite{zhang2015survey}~2015                  & 10       & 6       & 160                                                    & 50                                                    & D                                                   \\
OCCU~\cite{zhang2014evaluating}                & 5        & 5       & 30                                                     & 80                                                    & D                                                   \\
ACT42~\cite{cheng2012human}               & 24       & 14      & 48                                                     & 672                                                   & D, R                                              \\
Daily Living~\cite{zhang2012rgb}         & 5        & 5       & 10                                                     & 40                                                    & D, R, S                                           \\
NTU RGB+D~\cite{shahroudy2016ntu}             & 40       & 60      & 80                                                     & 4720                                                  & R, D, S, IR                                       \\
UWA3D~\cite{rahmani2014hopc}                 & 10       & 30      & 10                                                     & 290                                                   & R, D                                    \\         
\end{tabular}
\end{table}

Furthermore, genuine data may not always be readily available, particularly data of vulnerable people, due to complications in collecting and distributing it. The same applies for the data collection involving young people or children. Therefore, some AI applications focusing on assisting the everyday life of these target groups would likely be ineffective. There are ethical reasons which prohibit older people, people with disabilities and minors from participating in data collections that involve even daily activities, not to mention accidental falls i.e. for the development of fall detection algorithms \cite{xu2018new}. Similarly, data collection of abnormal behaviours, such as violent attacks \cite{mabrouk2018abnormal}, or of fighting actions for the development of a gaming application \cite{bloom2012g3d} is restricted due to health and safety risks and/or privacy protection issues. The few genuine data from actual scenes recorded in hospitals or assisted living homes is unfortunately not available for public distribution and research, mainly due to restrictions related to privacy and ethical approval policies. As a result, researchers have implemented human-simulated actions/scenarios in order to develop their detection algorithms and fill the data availability gap. Acting participants however, may not perform realistically an activity which in real life would have been performed in a spontaneous way, i.e. kicking and bouncing during a computer game, falling due to dizziness or stumbling on an obstacle, as their simulated movements would not be natural but rather mechanical.  Such implications make the data collection a difficult task as the actors can often find the tasks unpleasant or may feel unwilling to perform certain directions, depending on the risk factor of the tasks. The following sections discuss in detail the issues around existing datasets and recording practices and provide the reader with an insight into their limitations.

\subsection{Age of participants}
It is observed that most datasets provide limited data regarding the age of participants. In general, older people or minors are not represented in datasets - even if this data does not involve risk related actions. The available event data recordings are performed by mainly University students and researchers from an academic institution under instruction from the researcher to perform the necessary scenarios for data recording. In such circumstances, the ageing effect is not properly represented. In some other cases self-preservation may also take over, resulting in certain events being unrepresentative of genuine events or actions, particularly if the aim is to acquire data
representative of the vulnerable population (i.e. older people).

\subsection{Health of participants}
Participation in action recognition datasets normally involves an assessment of the actors' physical condition and/or mental state. If any issues are identified that would possibly be considered as a risk, such participants would be excluded due to restrictions set by ethics committees. Therefore, only the healthy subjects would participate in the data collection studies and vulnerable population would inevitably be excluded from these studies. 

\subsection{Variation of actions}

It is observed that there is a certain level of restriction of the acting freedom when performing tasks and therefore there is a small variation amongst repetitions. This is due to the fact that strict guidance is usually given to the subjects on how to perform. The example data in \cite{bloom2012g3d} shows how actions i.e. kicking and punching, visually appear similar between all subjects and within subjects themselves. Other datasets \cite{ma2014depth, kwolek2014human, gasparrini2016proposal, rahmani2014hopc} consist of samples from mainly one type of action performed in the same manner instead of including other possible variations of a particular action i.e. falling or sitting in a particular way. Another contributing factor to the limited variation is that when there is a  requirement for prompt data, time constrictions may not allow sufficient time for creating realistic scenarios of actions.  

\subsection{Size of datasets}
It is a known fact that the size of datasets for action recognition is generally limited. The small number of human actors performing events is usually not sufficient in order to represent the entire population. For example, one of the largest datasets \cite{shahroudy2016ntu} for action recognition consists of only 40 young, fit and healthy male and female subjects. Whilst, the number of participants may be sufficient for some detection tasks, it is not sufficiently varied. Especially in relation to the number of recorded real-life events (e.g. assaults, falls) in countries such as the US or the UK, the size of data samples captured for similar events is significantly smaller. The small number of samples can significantly affect the accuracy of a detection algorithm when applied to a real-life scenario, as the algorithm will have been trained on a small amount of data, thus limiting its performance.

\subsection{Variability of subjects}
Variability in human physical characteristics such as height, weight, age, or gender are factors which are generally ignored in data collection studies. However, an older person would normally have a different posture from a young person and a pregnant woman may walk differently from someone with a broken leg. Although it would not be feasible to ask subjects from all possible groups to perform actions, we would still need to consider the lack of variability as a limitation, as algorithms based on limited datasets would have questionable performance when applied to a broader demographic population. 

\subsection{Hesitation/avoidance/insouciance}

Human subjects performing staged actions may have difficulty in acting realistically due to hesitation associated with the concern of sustaining a possible injury. A \textit{hesitated action} is defined in this study as an event where the person tries to minimise the physical impact or possible harm of an action. In some circumstances, the instinct of self-preservation or lack of fitness may influence the action (e.g. kicking, punching, bouncing or falling) and therefore the event may be unrepresentative of a genuine one. As a result, data from these non-realistic recordings may have a negative impact on an algorithm's performance. The risk of injury is also an important factor when permission is sought to conduct experiments for action recognition of gaming applications (i.e. fighting) or assisted living (i.e. falls), hence, the type of actions may be conducted to follow a strict protocol specified by regulations of health and safety or ethical considerations. In these cases, researchers request disclaimers and inform the participants of their rights in the event of a sustained injury, particularly if they were deployed in a real environment. 


\subsection{Occlusions}
\begin{figure}
\centering
\includegraphics[width=0.5\textwidth]{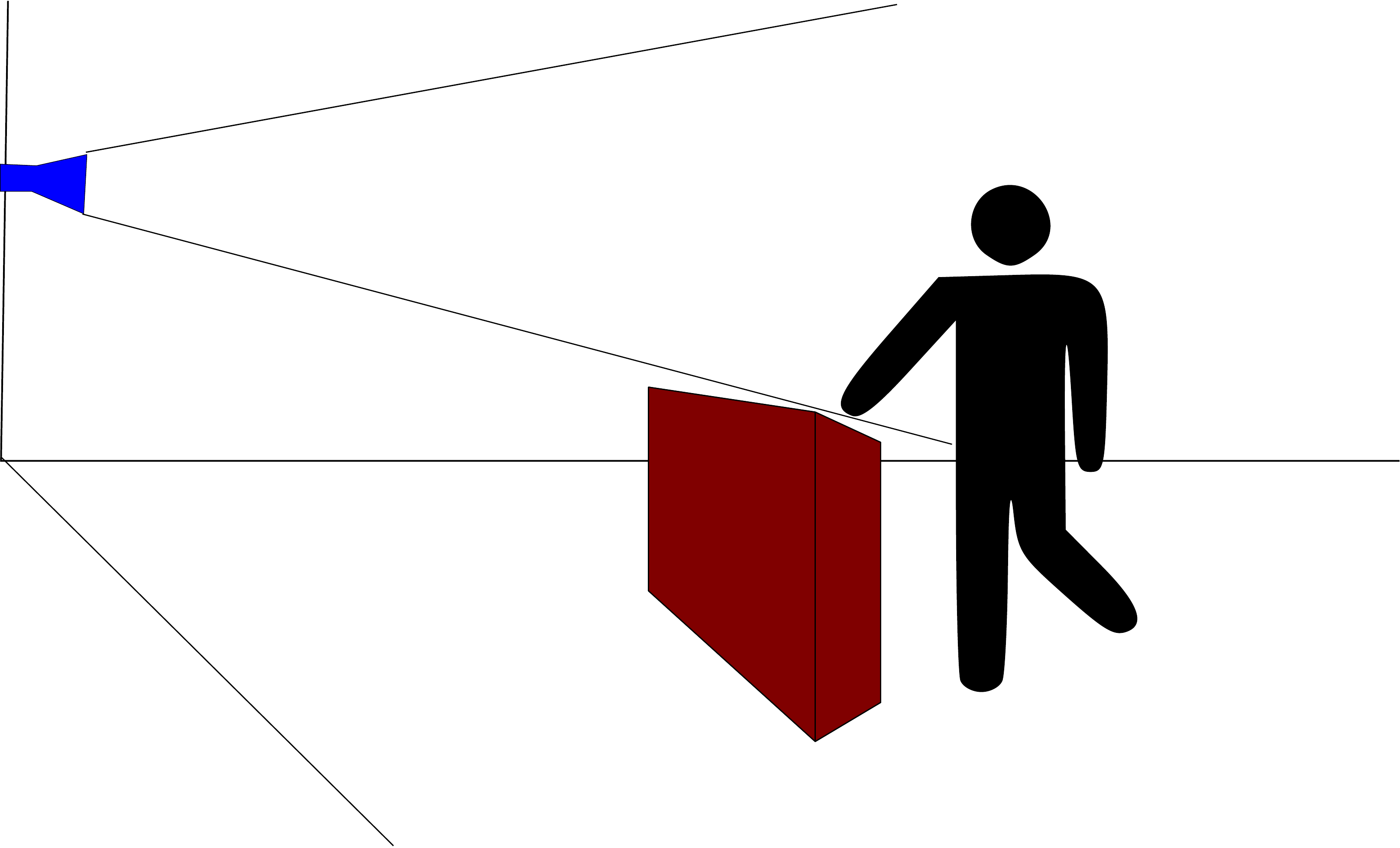}
\caption{Typical occluded scene. The camera view is partially blocked by the red box. Half the person is occluded.}
\label{fig:occlusionDiag}
\end{figure}

As already noted in the Introduction, studies mainly involve events that are generally appearing fully visible in the video scenes without any scene occlusions. Datasets generally include event videos without other objects appearing nearby unless this object is used, (i.e a chair, stool, or a bed) and as a consequence, occlusion scenarios are rarely represented. The lack of occlusions in most existing datasets offers an unrealistic perspective of virtually all indoor (i.e. home) environments. Therefore, in the event of an occluded action, current algorithms are generally untested for such scenarios. This could have devastating effects if a life threatening situation occurred when someone collapsed e.g. following a heart attack and the algorithm failed to recognise the action due to an occlusion. In a home scene we may sometimes get non-occluded views, but in reality people often  move around a cluttered environment and therefore, there may be frequent occasions during which they are part-occluded, to various degrees. Figure \ref{fig:occlusionDiag} illustrates an occlusion obstructing the view of a person. Although many studies discuss the application of detection algorithms at home or in hospital, occlusion is rarely mentioned, hence, methods are not evaluated to provide occlusion-robust solutions.

\subsection{Sensor location}
Only a few studies/datasets make note of the sensor location. The position of the sensor plays a significant role as to where the best field of view (f-o-v) is achieved in order to maintain a clear view of the home scene. This is unrelated to the minimisation of occlusions as even if the sensor is located higher, occlusions may still occur. The sensor location in some cases plays a significant role in how the person appears, hence, an algorithm is designed in order to detect an event using data from a particular viewpoint. See the example in \cite{kwolek2014human} where the depth sensor is located on the ceiling, pointing downwards. In other cases the sensor is placed on a table, which may not be the optimal location for a home scene. Obviously, the location in this case, aims to detect the height variation of i.e. a falling person, rather than how the depth or length of the human body changes during an event. Also, by placing the sensor at a low height, the view is more prone to self-occlusions.  In this scenario, an event may start near the sensor and conclude on the floor in front of the sensor and possibly under the f-o-v of the sensor - implying that the event is outside the viewing window. Other datasets are taken from the person's point of view (i.e. first person  \cite{pirsiavash2012detecting}). Such datasets may have other implications related to the camera motion, blurring, auto-focus delay, hand occlusions, lens distortion and limited field-of-view. Other datasets use videos from broadcast cameras and stationary/ PTZ CCTV where videos are provided and recorded continuously. In these cases, researchers would have to edit specific samples to match their requirements i.e. isolate the frames where a violent attack occurs or identify a free-kick movement during a sports game.

\subsection{Data quality and information adaptation}
One of the issues in using public datasets is the recording format and how other researchers can use the data. In some cases, depth data has been compressed resulting in poor depth information, or in other cases the depth information has been variant between the samples of the same dataset and therefore the reliability of this dataset has been questionable.  Also, different depth sensors technologies (e.g. Kinect, Orbbec, Primesense) or OpenNI/Microsoft Kinect SDK versions usually deliver different video/image formats which can be time-consuming to use or convert. 

In other cases, data requires cleaning (i.e images are blurred, oversaturated, noisy, out of focus etc.) and this task can take a significant amount of time in order to verify the processed material or to manually correct the imperfections. This implication is discussed in \cite{caba2015activitynet} where a large dataset was constructed using a crowd-sourcing approach for which participants were asked to perform and record several actions. The data adaptation and data cleaning in this study was a particularly time consuming task.

\subsection{Annotations}
In supervised learning, annotations are usually performed manually in order to name a particular event, action, motion, object, shape, colour etc. It is crucial that this step is performed with care, otherwise the developed approaches using this data will be inaccurate when tested. Other issues may arise, i.e. the machine learning algorithm may not converge during the training phase due to the misnaming of annotations (e.g. a banana is annotated as a melon and walking is annotated as running).

\subsection{Data disclosure}
In some cases data availability is limited due to privacy, copyright or intellectual property issues or there may be a cost in acquiring it. That is often the case with data being recorded in hospitals or personal homes and a special arrangement is required to be signed between the parties involved, which often comes with distribution restrictions to third parties.

\section{The issues of machine learning} 
The performance of the machine learning algorithms depends on the data used, the features and the classifier. The next subsections focus on these factors.

\subsection{Learning is as good as its data}

Recent developments in deep learning \cite{lecun2015deep} have increased the need for larger datasets for training. Nevertheless, the current visual data for computer vision methods as already discussed, is quite unrepresentative, particularly for action recognition. Therefore, a deep learning algorithm will ``suffer" in the same way as existing machine learning algorithms that are data-driven. Generally, it is known that a learning approach is as good as its data and this is discussed particularly in relation to deep learning approaches \cite{dlsid}. Hence, algorithms trained on limited datasets have questionable performance when applied to the wild. Furthermore, the process of cleaning and adapting the data to a specific format is done mainly manually. This requires a significant amount of time and effort and it would ultimately determine the algorithm's performance. Unfortunately, the use of unrepresentative or limited data has a negative impact on the performance as previously discussed in Section \ref{bigdataissues}.

\subsection{Features}
Selecting the right features for a particular task is crucial in all machine learning algorithms. Several selection algorithms have been developed to deliver the best possible feature or feature vector. Nevertheless, in supervised learning features are manually created by researchers who may not always know if these features are appropriate for the particular classification task. The process therefore depends on the human-like understanding of an event or action and is based on the researcher's perspective. The feature selection algorithms would select the best fit for the data it has been provided with, but this would be based on the initial set of features. This limitation cannot be easily overcome as even automatic feature generation studies \cite{katz2016explorekit} use existing human derived features to create new ones for supervised learning. To note, in unsupervised learning, several algorithms are reviewed in \cite{bengio2013representation} where authors discuss representation and deep learning algorithms for automatic feature extraction. The benefit of automatic extraction approaches is that an unsupervised algorithm will create its own representation of an object or event and potentially name it in its own language, which is effectively a step forward towards self-learning algorithms.

\subsection{Classifier - one-fits-all approach}

Another issue of data-driven algorithms is the lack of personalisation (one-fits-all), that is, a classifier unable to analyse different people's actions due to the variability of their physical characteristics or their manner of action. In a trainable algorithm, the requirement is to have a large enough dataset to capture natural variations of individual characteristics i.e height, weight, posture, fitness. This is not only to meet the data requirements of the machine learning algorithm but also to properly cover a range of people's physical characteristics and behavioural types, such as gait.

One example that supports this issue is found in \cite{GMPhDthesis} (Section 4.4) where a random search algorithm \cite{rastrigin1963convergence} failed to converge using a combined set of three datasets where human variability in physical characteristics and manner of action was different in each dataset. In short, two evaluation protocols where developed in order to use data from different datasets. These included a) a protocol where all samples from one dataset where used for training, then for testing on all samples of a different dataset, and b) a protocol where three datasets were combined and their samples where randomly selected for training and testing. A difficulty in convergence was identified when the subjects of the two different datasets (i.e. training and testing) had different physical characteristics and motion behaviour patterns. Also, the same issue was encountered when using a combined set of all datasets, where training did not always converge for the same reason. This is due to the fact that algorithms using a set of parameters for all data (i.e. of people with different physical characteristics) cannot address the variability within the sample and personalise the detection algorithms.

\section{Synthetic and simulation approaches}

With the rise of machine learning, the requirement of sufficiently large and variable datasets has become an issue, as such data may be laborious and expensive to acquire and label. One of the issues in many datasets and particularly the ones that involve visible facial characteristics or data from specific groups (i.e. older people, minors, people with disabilities, etc.) is that researchers cannot acquire such data from real events or that this data are not released due to privacy protection needs. One approach to deal with this problem is to generate data based on a combination of actual observations with simulation or synthetic data.

\begin{figure}
\centering
\includegraphics[width=0.8\textwidth]{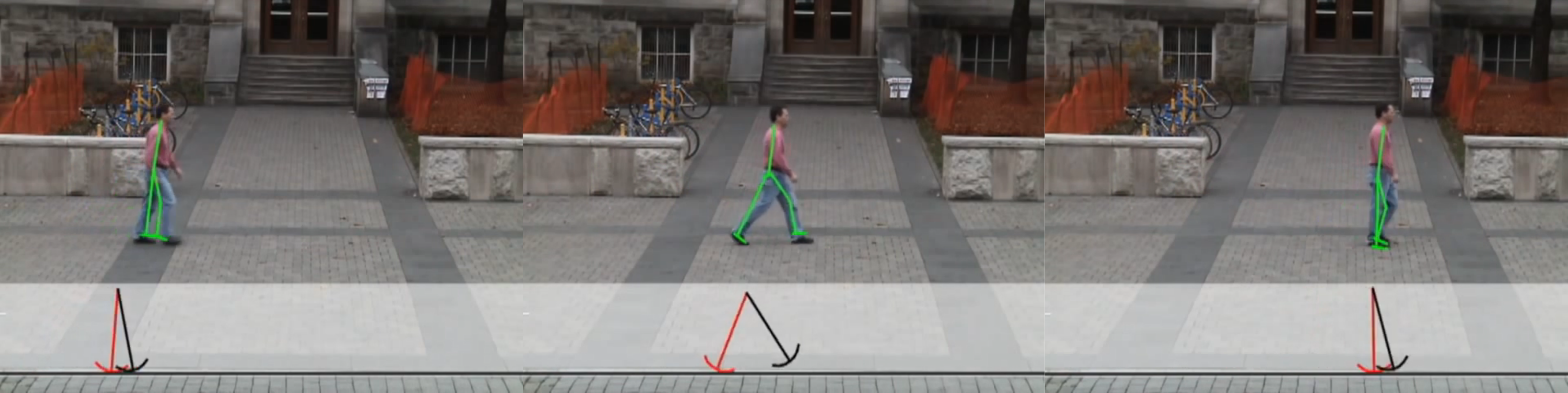}
\caption{Time lapse sequences of the Humanoid navigating different terrains \cite{heess2017emergence}}
\label{fig:brubaker}
\end{figure}

A number of studies employing computer vision and physics-based modelling exist in the literature. The most relevant studies \citep{xiang2010physics, brubaker2010physics} discuss how tracking a walking person can be achieved with the use of a bipedal model based on physics simulation. Brubaker et al. \cite{brubaker2010physics} discuss the use of a simple model for predicting the walking behaviour of a person. The authors evaluate their approach tested for varied walking speed including visual occlusion and also discuss the limitations of their approach and how a further complex model incorporating myoskeletal capabilities would provide a more accurate representation of human motion. Figure \ref{fig:brubaker} shows the motion correlation between the actual person walking and the stick model. 

Other studies propose physics-based frameworks for tracking articulated objects. In \citep{metaxas1993shape} Lagrange's equations of motion are used for models which can synthesise physically correct behaviours in response to applied forces and imposed constraints. Based on a previous study, the work in \citep{kakadiaris2000model} presents a mathematical formulation and implementation of a system capable of accurate general human motion modelling. The work in \citep{duff2011physical} uses an off-the-shelf physics simulator to track the behaviour of a rigid object. Another framework is presented in \citep{vondrak2012video}, where a method estimates human motion from monocular video. This is done by reconstructing three-dimensional controllers (i.e models) from the video which are capable of implicitly simulating the observed human behaviour. This behaviour is then replayed in other environments and under physical perturbations. Synthetic human data for activity monitoring is presented in \citep{zouba2007multi}. A dataset incorporating rigid poses is produced and used for the purpose of human behaviour recognition as well as scene understanding.

An example of synthetic data for action recognition can be found in the SOURREAL dataset presented in \citep{varol2017learning}, consisting of 6 million image frames together with ground truth pose, depth maps and segmentation masks. The amount of data is achieved by adding people's images of variable size as a foreground over a variety of background images. Other examples include synthetic datasets for pedestrian detection \cite{ekbatani2017synthetic} and synthetic urban scenes from the SYNTHIA dataset \cite{ros2016synthia}.

A bio-inspired modelling approach is presented in \cite{kazantzidis2018vide} where authors propose a novel video analysis paradigm ( i.e. ‘vide-omics’) for background / foreground segmentation. Inspired by the principles of genomics, this paradigm interprets videos as sets of temporal measurements of a scene in constant evolution without setting any constraint in terms of camera motion, object behaviour or scene structure. This puts variability at the core of every algorithm where the interpretation of scene mutations corresponds to video analysis. 

\begin{figure}
\centering
\includegraphics[width=0.7\textwidth]{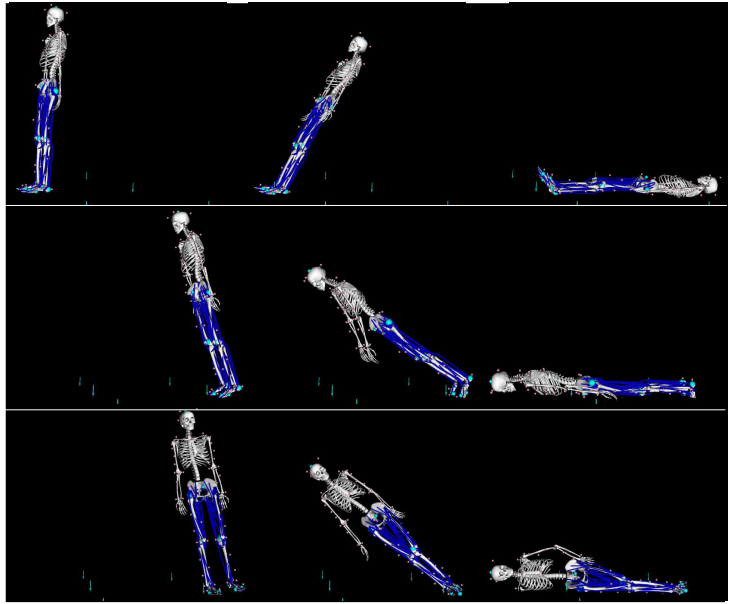}
\caption{Three types of rigid fall, backward (top), forward (middle) and sideways (bottom) as simulated on OpenSim}
\label{fig:all-fall-types-opensim}
\end{figure}

In \cite{mastorakis2018fall}, a different applicability of simulation is proposed. A myoskeletal simulation engine (OpenSim \cite{delp2007opensim}) is used to model several human falling events which are personalised based on the person's height. Other non-fall activities which have a similar to a fall behaviour i.e lying down, are also simulated and personalised. Their algorithm compares the simulation velocity profiles of falls and non-falls to determine whether an event is classified as a fall or not. The approach utilises a single fall and a non-fall modelling analysis and therefore, no machine learning is required for this method. This study outperforms data driven approaches when tested on public datasets.

\begin{figure}
\centering
\includegraphics[width=0.8\textwidth]{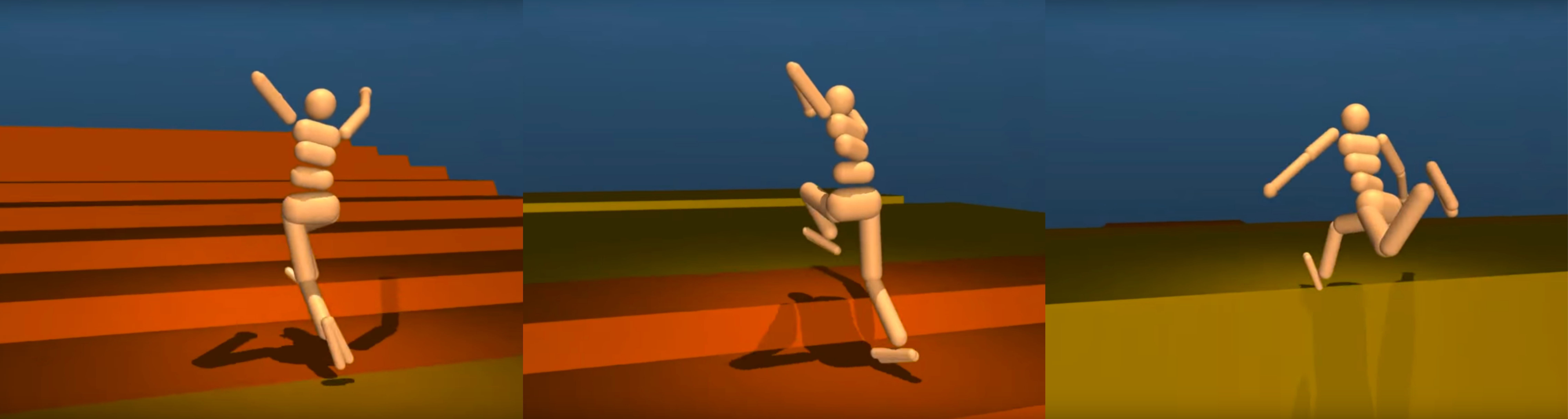}
\caption{Time lapse sequences of the Humanoid navigating different terrains \cite{heess2017emergence}}
\label{fig:deepmind}
\end{figure}

Apart from computer vision related studies, there are studies such as the work in \cite{li2016fall} which discusses the use of a physics-based simulation engine capable of detecting a rigid object's stability and likelihood of falling. Similarly, the study in \cite{lerer2016learning} presented how the behavior of simulated wooden blocks was used to train large convolutional network models which could accurately predict a collapsing outcome using an Intuitive Physics Engine  \cite{battaglia2013simulation}. Intuitive Physics is also discussed in \cite{kulkarni2016hierarchical} where authors explain in detail the human model of learning which develops with age.

In \cite{heess2017emergence} developed by DeepMind, several simulation actions were produced using a stick model performing activities such as running, jumping and climbing (see Figure \ref{fig:deepmind}) based on reinforcement learning. In this study, the algorithm performs self-learning training to overcome obstacles using Mujoco \cite{todorov2012mujoco} a physics simulation engine. In several released video examples, the stick model appears capable of learning how to run, showing in this way the advanced uses of a simulation based approach and reinforcement learning.

\begin{figure}
\centering
\includegraphics[width=0.8\textwidth]{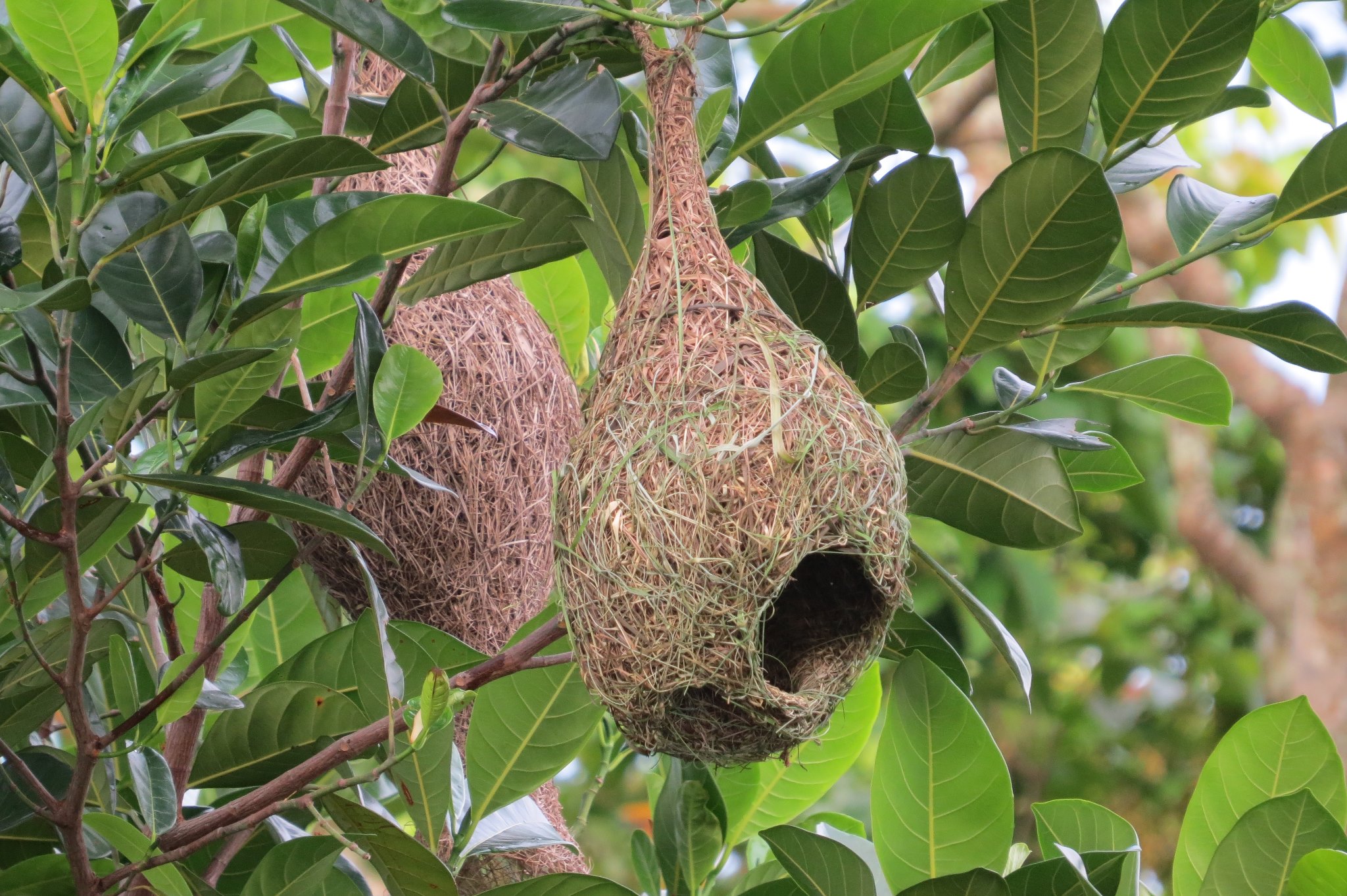}
\caption{Example of a weaverbird nest \cite{wiki:weaverbird}}
\label{fig:weaverbird}
\end{figure}

\section{Learning examples found in nature}
Weaverbirds are skillful animals capable of weaving complex structures as nests. A study \cite{bailey2015image} discusses the different patterns these birds apply in their weaving, pointing out the variability in patterns which is developed by learning. This learning process has developed by self-learning and self-correcting skills without observing other birds weaving their nests \cite{collias1964development}. It was also noted that birds change and improve their weaving techniques in their lifetime. However, the self learning approach is quite different in humans in relation to e.g. learning to speak or walk, actions that normally require tuition from someone else and are not self-taught. Furthermore, the self-learning bird knows how to improve itself without comparing its work with other birds. This may involve knowing how tightly or loosely to move the leaf fibers, grass, or twigs to make a strong nest. Effectively, the bird shows a certain level of understanding of physics i.e. it learns that a tight knot on a leaf fiber is more robust and the nest will keep for longer. The birds improve by making several nests in their lifetime, and therefore, self-correcting is a time related process.

\section{Discussion}

This paper addresses several complications around training machine learning algorithms. It is widely known, that data requirement is one of the factors that can have negative impact on the development of algorithms (see section 2). However, as several studies have shown, using synthetic data can overcome this issue. The use of synthetic visual data in studies has recently increased and several datasets \cite{gaidon2018reasonable} have been created. Such datasets contain new and more inclusive data that cover scene or action examples which are not available otherwise. In terms of specific examples of activities/events such as walking, falling people/objects, a simulation-based approach would resolve the data availability issue in terms of achieving realistic results and avoiding risks. Still, the question remains on how such data is evaluated and whether it fits the purpose. This would perhaps need to be further explored in order to fully ascertain the value of using synthetic data as seen in \cite{mastorakis2018fall} where the myoskeletal model was evaluated using YouTube videos. Another benefit of using synthetic approaches, is that they require less time in preparation as they do not involve human interaction for performing actions or scenarios. Instead, they would simulate the actions with a model, which is personalised using an individual's characteristics, such as their height, weight, posture etc. 

As seen in \cite{heess2017emergence}, the algorithm's self-learning capabilities are crucial for the task of walking as the system has no pre-knowledge of how walking behaviour is performed. Nevertheless, the behaviour of the stick model shows no improvement on the walking process. In other words, the algorithm does not know how to improve itself to walk as a human. One would say that the algorithm works instinctively, without having the ability to properly observe itself and therefore, the walking behaviour appears as rather unnatural. Self-correcting capabilities would be required in this approach but the algorithm fails to perform the correct way of walking, running or jumping and therefore, the model instead walks and jumps from one side to the other, in a quite unrealistic - for humans - way. Was the stick model suitable for this simulation or could a myoskeletal model have possibly performed better due to its human like physical characteristics?  Furthermore, the stick is modelled to stand ``a priory", i.e. it was not required to learn to stand, therefore, all actions start from the standing position, to the contrary of how humans learn to walk i.e. after they have learnt how to stand. In this case, walking is learned from the standing position which is used as pre-knowledge of upright posture that the algorithm tries to maintain throughout the actions. Due to these limitations, the algorithm does not appear to capture a natural human-like behaviour i.e. present a realistic walking/running/jumping action.

In other cases, self-learning capabilities are not required for modelling fall events as shown in \cite{mastorakis2018fall} where these events are performed by a myoskeletal model simulation in which, the behaviour of the simulation is affected by gravity and is parametrised by the person's height. In future studies, other parameters could also be used, such as posture, gait, weight etc. which could potentially alter the falling behaviour. These new parameters would allow the simulation models to perform more personalised types of falls and improve the non-machine learning approach described in Mastorakis et. al.

The example of a weaver-bird which can weave without copy-learning is an example of how self-learning and self-correction is performed in nature. This self-improvement mechanism, is a technique that has not been mastered in machine learning as yet. Currently, humans implement their own ``style" of learning to a machine which may not be the best possible pathway for successful and truly functional AI. Humans' cognitive evolution involved a successful taxonomy of plants and animals, yet we have not seen a successful example where these two classes are clustered accurately in unsupervised learning. The machine intelligence should ideally select its own valuable data - the one which (i.e. the machine) thinks is representative without human influence.  AI is still very much under the human supervision and would possibly require to evolve without the human influence in order to prove its potential and perform tasks without external learning i.e. without receiving tuition to practise a task. 

Humans can often recognise potential hazards via an understanding of physics, i.e. a wet surface may be slippery, not because they necessarily slipped or witnessed someone else slipping. This is because our brain has the ability to synthesise (i.e. simulate) information for awareness purposes. In a similar way, an algorithm would need to understand why a wet surface is slippery in order to adjust its behaviour. However, by implanting the knowledge that a particular wet surface is slippery we are possibly limiting the algorithm's ability to transfer its learning and adapt, as the algorithm is informed only on the condition (i.e. slippery) and not the reason that contributes to it (i.e. wet). Future AI algorithms could possibly utilise deep learning or other approaches to simulate and apprehend physics based events. An example of such an event would be what a machine can learn from a ball's behaviour when it simulates a ball throwing, knowing its initial velocity. Would the machine need to practise the simulation in order to learn to throw the ball back to the original height? If a machine would develop the ability to understand the effects of physics such as, the impact of gravity, it could then be capable of controlling the learning process without the use of a physics engine or video data.

\section{Conclusion}
This study discusses the current issues of data-driven approaches in computer vision and particularly in action recognition, where video data is required for processing. It was highlighted that current datasets are rather limited and unrepresentative in terms of variability in physical characteristics and patterns of behaviour as well as due to issues around scene setup, occlusions, data adaptation and privacy, amongst others. Certain studies developing synthetic/simulation data to overcome the data availability issue are also discussed. It was observed that these datasets increased the number of samples, although they did not necessarily enhance the variability of events, scenes or scenarios.

Furthermore, non-data-driven modelling approaches were presented (i.e. for walking/falling and background subtraction) showing the current proof-of-concept, which would require however, further development in order to generalise their applicability. We have seen that the use of a simulation engine may not always be beneficial as the models used are pre-built and have their own limitations and constraints. Therefore, an ideal learning approach would involve the algorithm learning and correcting itself in order to improve and perform in a flawless manner without the need to observe how humans perform a similar task. Self-learning approaches would benefit from the understanding of physics or other natural sciences in order to improve their performance and develop self-correction abilities. The machine based self-correction abilities would then outperform the human ones and lead the path to advanced artificial intelligence systems.



\bibliographystyle{elsarticle-num-names} 
\bibliography{AI.bib}





\end{document}